\documentclass[11pt]{article}
\usepackage[utf8]{inputenc}
\usepackage[margin=1in]{geometry}

\usepackage{microtype}
\usepackage{graphicx}
\usepackage{subfig}
\usepackage{booktabs} 
\usepackage{natbib}
\setcitestyle{open={(},close={)}}
\usepackage[colorlinks=true,citecolor=blue]{hyperref}

\usepackage{graphicx}

\usepackage{tikz}
\usepackage{comment}
\usepackage{amsmath,amssymb} 
\usepackage{color}

\usepackage[accsupp]{axessibility}  

\usepackage{graphicx}
\usepackage{amsmath}
\usepackage{amssymb}
\usepackage{booktabs}

\usepackage{caption}
\usepackage{soul}

\usepackage{multirow}
\usepackage{enumitem}

\usepackage{algpseudocode}
\usepackage{algorithm}

\usepackage{wrapfig,lipsum}

\begin{document}

\title{\bf\Huge Defending Backdoor Attacks on Vision Transformer via Patch Processing}

\author{\vspace{0.5in}\\\textbf{Khoa D. Doan, Yingjie Lao, Peng Yang, Ping Li} \\\\
Cognitive Computing Lab\\
Baidu Research\\
10900 NE 8th St. Bellevue, WA 98004, USA\\\\
  \texttt{\{khoadoan106, laoyingjie, pengyang5612, pingli98\}@gmail.com}
}

\date{\vspace{0.5in}}
\maketitle

\begin{abstract}\vspace{0.3in}

\noindent\footnote{The paper will appear in the Proceedings of the AAAI'23 Conference. This work was initially submitted in November 2021 to  CVPR'22, then it was re-submitted to ECCV'22. The paper was made public in June 2022. The authors sincerely thank all the referees from the Program Committees of CVPR'22, ECCV'22, and AAAI'23.}Vision Transformers (ViTs) have a radically different architecture with significantly less inductive bias than Convolutional Neural Networks. Along with the improvement in performance, security and robustness of ViTs are also of great importance to study. In contrast to many recent works that exploit the robustness of ViTs against adversarial examples, this paper investigates a representative causative attack, i.e., backdoor. We first examine the vulnerability of ViTs against various backdoor attacks and find that ViTs are also quite vulnerable to existing  attacks. However, we observe that the clean-data accuracy and backdoor attack success rate of ViTs respond distinctively to patch transformations before the positional encoding. Then, based on this finding, we propose an effective method for ViTs to defend both patch-based and blending-based trigger backdoor attacks via patch processing. The performances are evaluated on several benchmark datasets, including CIFAR10, GTSRB, and TinyImageNet, which show the proposed novel defense is very successful in mitigating backdoor attacks for ViTs. To the best of our knowledge, this paper presents the first defensive strategy that utilizes a unique characteristic of ViTs against backdoor attacks.
\end{abstract}

\newpage

\section{Introduction}
\label{sec:intro}
The versatility of machine learning makes it a promising technology for implementing a wide variety of complex systems such as autonomous driving~\citep{grigorescu2020survey,caesar2020nuscenes}, intrusion detection~\citep{vinayakumar2019deep,berman2019survey,doan2021interpretable}, communication~\citep{huang2019deep}, pandemic mitigation~\citep{oh2020deep,alimadadi2020artificial}, retrieval~\citep{doan2022one,doan2020imagehashing}, and targeted advertising~\citep{doan2019adversarial} systems. These examples also illustrate that a large portion of safety-critical applications are benefited from the evolution of machine learning, which meanwhile requires high degrees of security and trustworthiness of these technologies~\citep{yang2021robust,lao2022identification,lao2022deepauth,zhao2022integrity}. Unfortunately, vulnerabilities have emerged from many aspects of machine learning and a wide body of research has been investigated recently to exploit both these vulnerabilities and defensive measures to mitigate attacks against machine learning, especially for deep learning systems~\citep{szegedy2013intriguing,liu2018survey,akhtar2018threat}. 

One such vulnerability, backdoor attack, allows an adversary with access to the model's training phase the possibility of injecting backdoors to maliciously alter the machine learning model behavior~\citep{liu2017trojaning,chen2017targeted}. These backdoor injection attacks poison the training data or modify the learning algorithm such that an association between a specific adversarial input ``trigger'' and an adversarial output ``behavior'' is formed. A trigger is typically locally superimposed on a clean image with an image pattern (i.e., patch-based)~\citep{gu2019badnets,liu2017trojaning} or globally blended (i.e., blending-based)~\citep{barni2019new,liu2020reflection,nguyen2021wanet,doan2021backdoor,doan2021lira,doan2022marksman} for improving the stealthiness. The compromised model will continue to behave normally as intended under the typical usage scenarios with clean inputs. But by exposing the model to the correct triggers, a user with the prerequisite knowledge can then directly control the model's prediction.

As machine learning continues to improve upon its current success, developers must understand both the vulnerabilities that machine learning brings and valid methods for overcoming these weaknesses. One recent major advance in computer vision tasks is the vision transformer (ViT)~\citep{dosovitskiy2020image}, which adapts the multi-head self-attention mechanism from the natural language processing (NLP) tasks. Specifically, during ViT's training, images are pre-processed as patches, which are treated similarly to words in NLP. It has been shown that ViT can achieve comparable or even better performance to state-of-the-art convolutional neural network (CNN) architectures on various vision tasks~\citep{dosovitskiy2020image,liu2021swin,wang2021pyramid,touvron2021training,yuan2021tokens,gkelios2021investigating,khan2021transformers,chen2021mvt,yu2022boat,yu2022degenerate}. 

While switching from convolution to self-attention has shown promising outcomes in tackling these vision tasks from the performance perspective, the implications of such fundamental differences on security and robustness are also of paramount importance to study. Several recent works examined the performance of ViT against adversarial examples~\citep{mao2021towards,benz2021adversarial,bhojanapalli2021understanding,naseer2021intriguing,mahmood2021robustness,shao2021adversarial,naseer2022improving,joshi2021adversarial}. However, the vulnerability of ViT against backdoor attacks and the corresponding countermeasures have not been extensively studied. In fact, to the best of our knowledge, only one very recent work looked at this direction~\citep{lv2021dbia}, which proposed a data-free backdoor embedding attack against the vision transformer networks. In contrast to this prior work, we focus on the defensive side. Aligning with the processing of ViT that divides an image into patches, we mainly study the implications of patch transformations on image classification tasks in this paper. Specifically, we utilize two techniques, namely PatchDrop and PatchShuffle, which randomly drop and shuffle patches of an image, respectively. Under these patch processing, we find that ViT exhibits a different characteristic from CNN models and also responds distinctively between clean samples and backdoor samples. Specifically, PatchDrop is effective in detecting patch-based backdoor attacks, while PatchShuffle can successfully mitigate blending-based backdoor attacks. Therefore, based on patch processing, we propose a novel defensive solution to combat backdoor attacks. 

The contributions of this paper are summarized below: 
\begin{itemize}
    \item We first perform an empirical study on the vulnerability of ViTs against both patch-based and blending-based backdoor attacks and find ViTs are still quite vulnerable to backdoor attacks.
    \item We observe an interesting characteristic of ViTs that clean-data accuracy and backdoor attack success rate of ViTs respond distinctively to patch processing before the positional encoding, which is not seen on CNN models.
    \item We propose a novel defensive solution to mitigate backdoor attacks on ViTs via patch processing. We analyze two processing methods, i.e., PatchDrop and PatchShuffle, and examine their effectiveness in reducing the attack success rate (ASR) of backdoor attacks. In particular, PatchDrop and PatchShuffle are effective in detecting patch-based and blending-based backdoor attacks, respectively. Together, they are used to effectively detect the backdoor samples without prior knowledge of whether the attack is patch-based or blending-based.
    \item We comprehensively evaluate the performance of the proposed techniques on a wide range of benchmark settings, including CIFAR10, GTSRB, and TinyImageNet. 
\end{itemize}

\vspace{0.1in}
\section{Related Work}

Previous works on deep neural network (DNN) backdoor injection have understood the attack as the process of introducing malicious modifications to a model, $F(\cdot)$, trained to classify the dataset ($X$,$Y$). These changes force an association with specific input triggers, ($\Delta$, $m$), to the desired model output, $y_t$~\citep{gu2019badnets,liu2017trojaning,bagdasaryan2021blind,yao2019latent}. Through Equation~(\ref{eq:1}), the trigger can be superimposed on any input such that a poisoned input is formed.
\begin{equation}
    P(x,m,\Delta) = x \circ (1-m) + \Delta \circ m 
    \label{eq:1}
\end{equation}
Here we use $\circ$ to denote the element-wise product and $m$ is a mask used to determine the region of the input containing the trigger pattern, $\Delta$. In essence, the adversarial goal is to force the model to minimize the compound loss function: $\mathcal{M}(F_{\omega}(x),y) + c \cdot D(F(P(x,m,\Delta)),F(x_t))$, instead of the original benign loss function such as cross-entropy loss,  where $D(\cdot,\cdot)$ defines the similarity between the model's actual behavior and a target behavior described by the input $x_t$ while the constant $c$ is used to balance the terms~\citep{yao2019latent}.

The main methodologies used to inject this functionality into the model are contaminating the training data~\citep{chen2017targeted,liu2017trojaning,gu2019badnets,saha2020hidden}, altering the training algorithm~\citep{bagdasaryan2021blind} or overwriting/retraining the model parameters after deployment~\citep{dumford2018backdooring}. Besides the original patch-based trigger~\citep{gu2019badnets}, various blending-based trigger patterns have also been proposed, including blended~\citep{chen2017targeted}, sinusoidal strips (SIG)~\citep{barni2019new}, reflection (ReFool)~\citep{liu2020reflection}, and warping (WaNet)~\citep{nguyen2021wanet}. Note that in order to differentiate from the patch used in describing the processing of ViTs, we limit the usage of patch for backdoor attacks to only ``patch-based''. In other words, \textbf{only ``patch-based'' refers to the backdoor attack, while all the other usages of ``patch'' are related to the ViTs in this paper}. For the backdoor embedding attack on ViT~\citep{lv2021dbia}, it seeks to catch most attention of the victim model by leveraging the unique attention mechanism. 

On the other hand, several categories of defensive solutions have also been proposed to combat backdoor attacks in past years~\citep{chen2018detecting,tran2018spectral,gao2019strip,liu2017neural,li2020rethinking,liu2018fine,cheng2020defending,wang2019neural,chen2019deepinspect,qiao2019defending}. One direction is to remove, detect, or mismatch the trigger of inputs through certain processing or transformations of the input images~\citep{liu2017neural,li2020rethinking,doan2020februus,udeshi2019model,qiu2021deepsweep,gao2019strip}. Note that most of these defensive methods are model-agnostic and mainly target at processing the inputs. Our proposed defensive method follows a similar concept as these input processing methods. For instance, similar to STRIP~\citep{gao2019strip} that examines the entropy in predicted classes after a set of input perturbations to check any violation of the input-dependence property of a benign model, we leverage the distinctive performance between the clean sample and backdoor sample against patch processing to detect malicious behaviors. Another advantage of such methods, including the proposed one, is that they only require access to clean samples, which is a more practical setting for defending backdoor attacks.

\section{Backdoor Attacks on ViT} \label{test_attack}

\subsection{Threat Model}
We follow the typical threat model of DNN backdoor attacks~\citep{gu2019badnets} that a user wishes to establish a model for a specific image classification task by training with data provided by a third party. We assume the adversary has the capability of injecting poisoned data samples into the training dataset, but cannot modify the model architecture, the training setting, or the inference pipeline. Since the user will check the accuracy of the trained model on a held-out validation dataset (clean samples), the adversarial goal is to embed a backdoor into the model through data contamination without degrading the clean-data accuracy over the image classification task. In other words, the model should produce malicious behavior only on images with the trigger for the backdoor, while performing normally otherwise. 

\subsection{Attack Experimental Results}

To understand the security threat on ViTs against the backdoor attacks, we consider two most popular approaches of creating the backdoor triggers: local patch-based triggers, BadNets~\citep{gu2019badnets} and SinglePixel~\citep{bagdasaryan2021blind}, and global blending-based trigger, ReFool~\citep{liu2020reflection} and WaNet~\citep{nguyen2021wanet}. We evaluate the performance on CIFAR10, GTSRB, and TinyImageNet datasets.

Specifically, we perform the attack experiment by poisoning the training dataset and the corresponding ground-truth labels. For each training dataset, similar to prior works~\citep{gu2019badnets,nguyen2021wanet,liu2020reflection}, we select a small number of samples (less than 10\%) and apply the corresponding trigger on each of the selected images.

\newpage

\begin{figure}[t]
	\centering
		\subfloat[GTSRB/BadNets]{\includegraphics[width=4.2in]{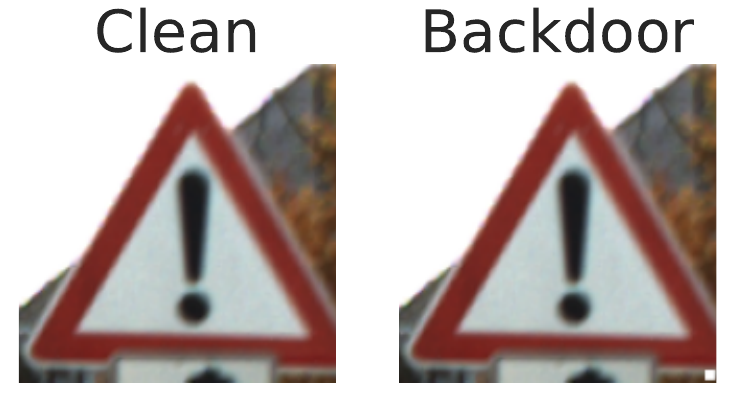}}\\
		\subfloat[TinyImageNet/ReFool]{\includegraphics[width=4.2in]{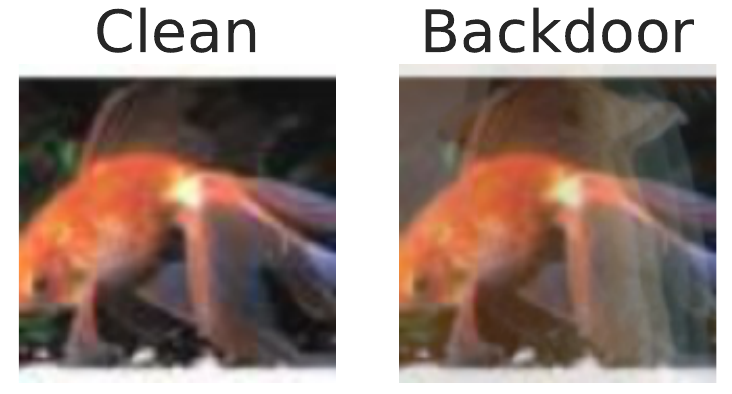}}
	\caption{Clean and backdoor samples with local patch-based trigger (a square in bottom right corner) and global blending-based trigger (an embedded reflection). }\label{fig:sample_patch}
\end{figure}

Figure~\ref{fig:sample_patch} shows some examples of both patch-based and blending-based backdoor samples. The labels of the poisoned samples are also changed to the target label. The poisoned training data are then used to train the image classification model. Then, we perform training using two ViT variants, the original ViT~\citep{dosovitskiy2020image} and DeiT~\citep{touvron2021training}, and several other popular CNN model architectures, including Vgg11~\citep{simonyan2014very}, ResNet18~\citep{he2016deep}, and Big Transfer (BiT)~\citep{kolesnikov2020big}. Note that the models are pre-trained on ImageNet-21k and fine-tuned on the corresponding dataset to ensure a consistent experimentation framework. This setup is influenced by the fact that large-scale ViTs and BiT are not trained from scratch on smaller-scale datasets to prevent overfitting. The detailed training settings, statistics of datasets, and model architectures are summarized in the supplementary materials. Each trained model is then evaluated on the held-out test sets of clean and backdoor samples. The backdoor samples are applied with the triggers that are generated using the same mechanism in the corresponding attack strategy for the evaluation. 

\newpage

In Table~\ref{tab:attack_patch_based} and Table~\ref{tab:attack_global_trigger}, we show the clean-data and backdoor-data performance of the trained models for BadNets and WaNet, respectively. We can observe that the trained ViT and DeiT with the backdoors have similar, high clean-data accuracies to that of the corresponding benign models (still outperforming other CNN models). However, when the triggers are present, the probabilities of the poisoned ViT models to predict the target label (i.e., ASR) are also quite high, which are above 96\% on all datasets. In other words, ViTs are at least as vulnerable against backdoor attacks as the CNN models. In fact, the patch-based backdoor attack on ViTs seems to be even slightly more successful than on other CNN models, which further validates the need for studying the backdoor attacks and countermeasures on ViTs. We observe similar results for SinglePixel and ReFool attacks.

\begin{table*}[t!]
    \centering
     \caption{Patch-based Backdoor Attack (BadNets)}
     \resizebox{\columnwidth}{!}{
    \begin{tabular}{c|c|c|c|c|c|c|c|c|c|c}
        \toprule
        \multirow{2}{*}{Dataset} & \multicolumn{2}{c}{ViT} & \multicolumn{2}{|c}{DeiT} &
        \multicolumn{2}{|c}{Vgg11} & \multicolumn{2}{|c}{ResNet18} & \multicolumn{2}{|c}{BiT}\\ \cline{2-11}
          & Clean & Attack & Clean & Attack & Clean & Attack & Clean & Attack & Clean & Attack\\ \hline
        CIFAR10 & 98.93  & 98.47 & 98.82  & 97.82  & 93.44 & 96.95  & 96.77 & 97.17 & 98.51 & 97.09 \\
        GTSRB & 98.68  & 96.46 & 98.55  & 95.62 & 98.05 & 91.21 & 98.86 & 94.33 & 98.71 & 94.77 \\
        TinyImageNet & 86.46  & 98.02 &  87.76  & 95.77 & 61.94 & 88.57 & 68.67 & 96.16 & 80.99 & 96.94 \\ \bottomrule
    \end{tabular}
    }
    \label{tab:attack_patch_based}
\end{table*}

\begin{table*}[t!]
    \centering
    \caption{Blending-based Backdoor Attack (WaNet)}
    \resizebox{\columnwidth}{!}{
    \begin{tabular}{c|c|c|c|c|c|c|c|c|c|c}
        \toprule
        \multirow{2}{*}{Dataset} & \multicolumn{2}{c}{ViT} & \multicolumn{2}{|c}{DeiT} &
        \multicolumn{2}{|c}{Vgg11} & \multicolumn{2}{|c}{ResNet18} & \multicolumn{2}{|c}{BiT}\\ \cline{2-11}
          & Clean & Attack & Clean & Attack & Clean & Attack & Clean & Attack & Clean & Attack\\ \hline
        CIFAR10  & 97.88  & 99.98  & 97.92 & 99.99  & 95.06 & 99.71 & 96.31  & 99.92 & 97.85 & 99.99 \\
        GTSRB & 99.08  & 99.74 & 97.55  & 98.27 & 98.75 & 99.48 & 99.28 & 99.74 & 99.23 & 99.98 \\
        TinyImageNet & 77.48  & 99.99 &  83.90  & 98.53 & 64.96 & 99.21 & 68.23 & 99.98 & 75.90 & 99.99 \\ \bottomrule
    \end{tabular}
    }
    \label{tab:attack_global_trigger}
\end{table*}

\section{Backdoor Attacks vs. Patch Processing} 

In the previous section, we have shown that backdoor attacks are still quite successful on ViTs. Besides, as we discussed above, it has also recently been shown that ViTs are vulnerable against other types of attacks, although they exhibit certain degrees of improvement in robustness against the transferability of adversarial examples~\citep{mahmood2021robustness,shao2021adversarial}. While these features of ViTs are similar to the CNN models, ViTs have also been shown to be more robust toward occlusions, distributional shifts, and permutation~\citep{naseer2021intriguing}. In this section, we extend the robustness study of the receptive fields of ViTs with respect to the backdoor attack models and compare their performance to CNN models. 

\subsection{Patch Processing}

Following the existing defensive methods that process images at the input space for detecting backdoor attacks~\citep{liu2017neural,li2020rethinking,doan2020februus,udeshi2019model,qiu2021deepsweep,gao2019strip}, we study the performance of the backdoor attacks on ViT models through input transformations that align with the characteristic of ViTs, i.e., patch processing where the content of the image is randomly perturbed. Specifically, each input image $x$ is represented as a sequence of patches with $L$ elements: $\{x_i\}_{i=1,..,L}$. Note that the patch $x_i$ does not necessarily have the same size as the patch size used in the pre-trained ViT model. Perturbing the image's patches is equivalent to modifying its content. Here, we focus on the question: \textit{How does perturbation influence the receptive field of ViTs on image patches when various backdoor triggers are present}? 

We denote the patch processing on $x$ with a function $R$ and consider the following strategies for performing the patch processing:
\begin{itemize}
    \item \textbf{PatchDrop.} Similar to~\citet{naseer2021intriguing}, we randomly drop $M$ patches from the total $L$ patches of an image $x$. We divide the image into $L=l \times l$ patches that belong to a spatial grid of $l \times l$. The number of dropped patches indicates the information loss on the image content.
    \item \textbf{PatchShuffle.} We randomly shuffle the $L$ patches of an image $x$. The $L$ patches are created in similar spatial grids as those of PatchDrop. PatchShuffle does not remove the content of the image but can significantly impact the receptive fields of the models. 
\end{itemize}

Note that similar forms of patch transformations on ViT models have been considered in prior works~\citep{naseer2021intriguing,shao2021adversarial}, but not in the context of backdoor attacks. 

\subsection{Performance of Backdoor Attacks against Patch Processing}

\begin{figure}[b!]

{\hspace{-0.25in}
		\subfloat[CIFAR10]{\includegraphics[width=1.05 \textwidth]{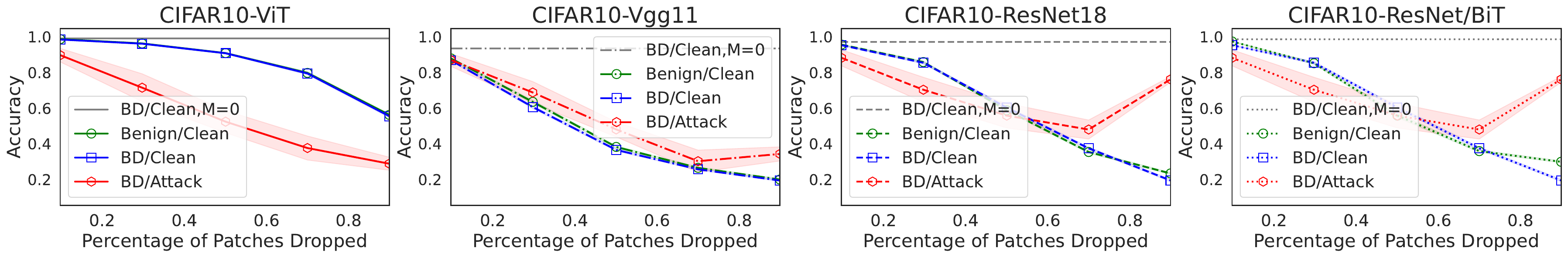}}
}

{\hspace{-0.25in}
		\subfloat[GTSRB]{\includegraphics[width=1.05 \textwidth]{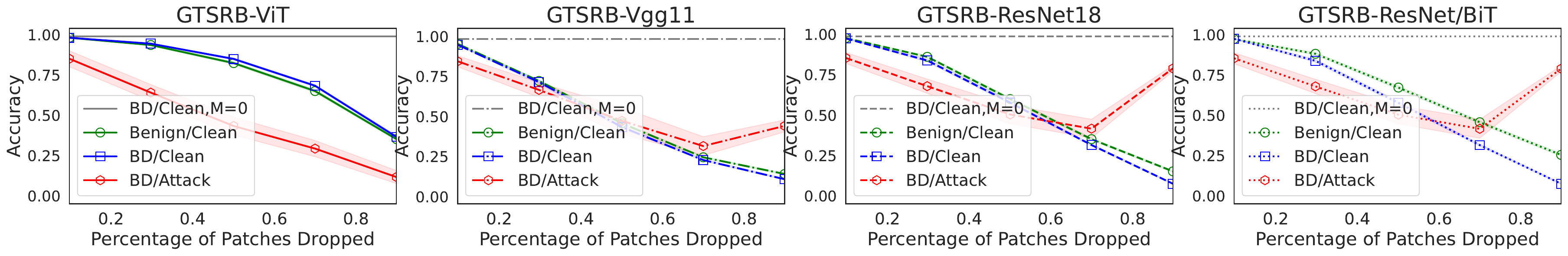}}
}

{\hspace{-0.25in}
		\subfloat[TinyImageNet]{\includegraphics[width=1.05 \textwidth]{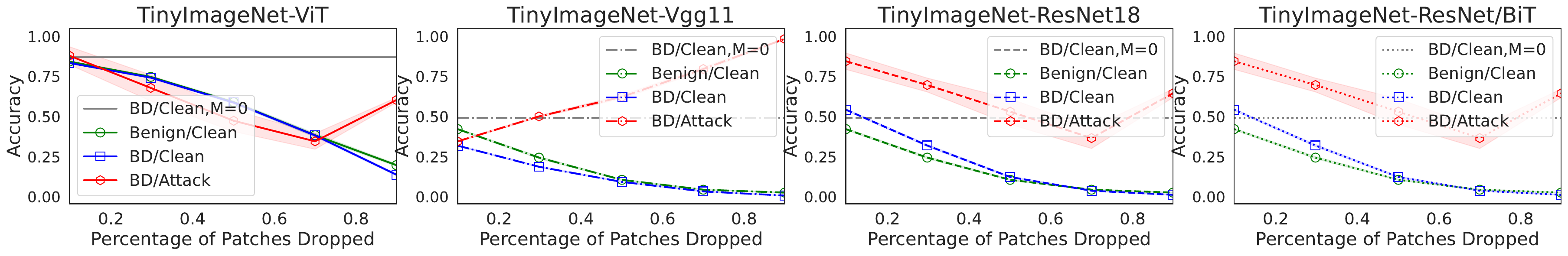}}
}	
	\caption{Performance of clean-data accuracy and backdoor attacks with dropped patches on ViT, Vgg11, ResNet18, and BiT.}\label{fig:result_patch_drops_acc}
\end{figure}

\begin{figure}[!htpb]
{\hspace{-0.25in}
\subfloat[CIFAR10]{\includegraphics[width=1.05 \textwidth]{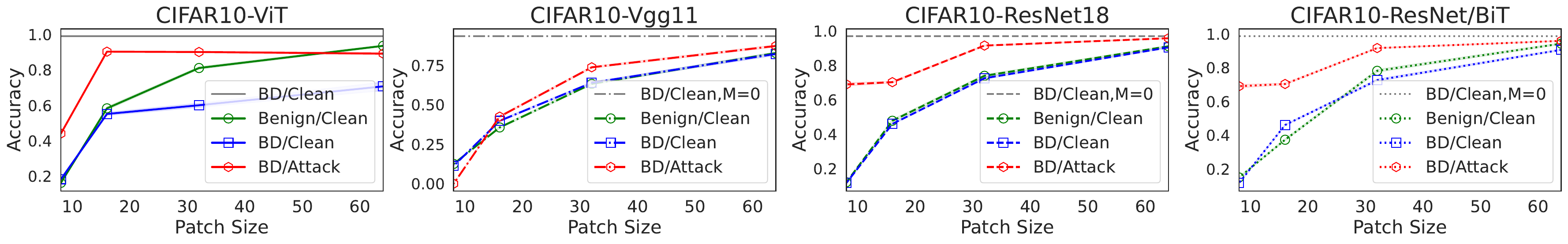}}
}

{\hspace{-0.25in}
		\subfloat[GTSRB]{\includegraphics[width=1.05 \textwidth]{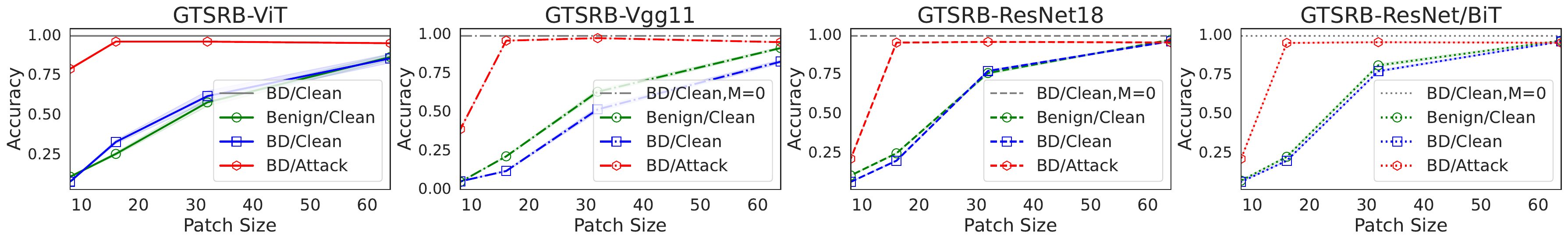}}
}

{\hspace{-0.25in}
		\subfloat[TinyImageNet]{\includegraphics[width=1.05 \textwidth]{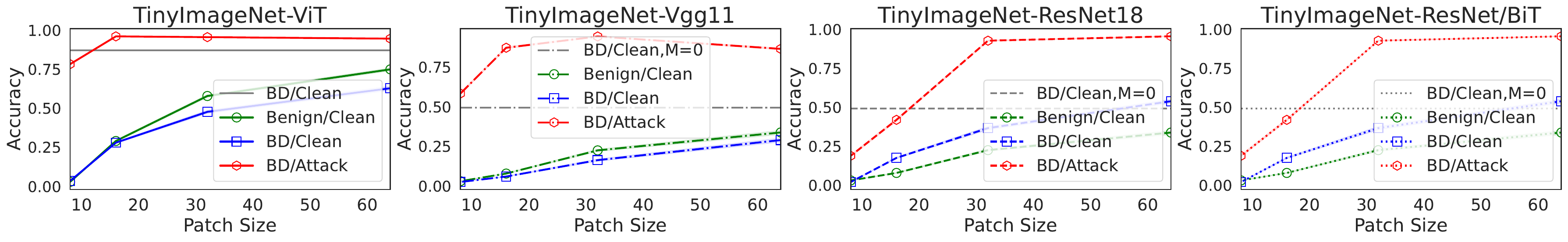}}
}
	\caption{Performance of clean-data accuracy and backdoor attacks with dropped patches on ViT, Vgg11, ResNet18, and BiT.}\label{fig:result_patch_drops_acc_refool}\vspace{0.2in}
\end{figure}

Similar to the attack experiments, we evaluate the performance on three datasets. We first empirically study the trends of backdoor ASR and clean-data accuracy with respect to patch processing on the corresponding test set for each dataset. We measure the clean-data performance (i.e., test accuracy) and the backdoor performance (i.e., ASR) for each experiment. The results are reported in Figure~\ref{fig:result_patch_drops_acc} and Figure~\ref{fig:result_patch_drops_acc_refool} for BadNets and ReFool, respectively. 

\newpage

For patch-based attacks with PatchDrop, we observe that the clean-data performances of ViT only drop slightly on CIFAR10 and GTSRB even when almost 50\% of the image content is removed. In contrast, the clean-data performances drop much more significantly in all the other three CNN models. On TinyImageNet, the clean-data performance of ViT drops more than in the other datasets. However, when the backdoor triggers are present, the attack success rate on ViT decreases significantly, even with a slight loss in the content of the images. In comparison, backdoor attacks on CNN models are more robust to PatchDrop. Interestingly, if we continue to drop more patches, the ASR on the CNN models suddenly increases in several experiments. A possible explanation is that CNN models and backdoor attacks rely on smaller regions of the image than ViT for prediction and achieving the target classes, respectively, which makes the clean-data accuracy of ViT more robust to patch processing. We also notice another important result: the variance in the predictions of the poisoned models is higher for backdoor samples than for the clean samples. We summarize the observations for patch-based attacks with respect to PatchDrop as follows:
\begin{itemize}
    \item Clean-data accuracy sensitivity: ViT $<<$ CNN
    \item ASR sensitivity: ViT $>$ CNN
    \item \textbf{Gap between accuracy and ASR: ViT} $>$ \textbf{CNN}
\end{itemize}


\newpage

However, for blending-based attacks with PatchDrop, we do not observe a consistent difference between the ViTs and CNN models, although ViTs are more robust with respect to clean-data accuracy and ASR. Since the trigger is well-blended into the images across the entire pixel space, as in ReFool and WaNet, PatchDrop tends to be less impactful on the backdoor, similar to the robustness of the models on the foreground objects. However, for blending-based attacks with PatchShuffle, we observe that the clean-data performances of ViT drop significantly. In contrast, the ASRs only drop slightly. Such robustness of the trigger is consistent across various patch sizes (i.e., $|x_i|$) on all datasets. For the CNN models, the gaps between clean-data accuracy and ASR are smaller; in some cases, e.g., Vgg11, the gap can become significantly narrow. In previous studies, ViTs exhibit high robustness against patch transformation for larger patch sizes~\citep{naseer2021intriguing}. Under our patch processing technique PatchShuffle, the significant robustness of the trigger across all patch sizes, especially the smaller sizes, is interesting. Such performance can possibly be explained that ViTs learn and generalize the spatial invariance of the triggers extremely well. Similarly, we summarize the observations for blending-based attacks with respect to  PatchShuffle as follows: 
\begin{itemize}
    \item Clean-data accuracy sensitivity: ViT $>$ CNN
    \item ASR sensitivity: ViT $<<$ CNN
    \item \textbf{Gap between accuracy and ASR: ViT} $>$ \textbf{CNN} 
\end{itemize}

\vspace{0.2in}

In summary, ViT has distinguishable performance between clean-data performance and ASR against certain patch processing techniques: the ASR drops significantly on ViT for \textbf{patch-based attacks with PatchDrop}, while the clean-data performance drops significantly on ViT for \textbf{blending-based trigger attacks with PatchShuffle}. As a result, for both cases, ViT has a larger gap between accuracy and ASR than CNN. It is important to note that these characteristics are not observed on CNN models. Therefore, the observed impact of patch processing against backdoor attacks is unique to ViT. 

\section{Novel Defensive Solution for ViT} \label{vit_defense}
\subsection{Methodology}

Based on our observations above, we propose an effective backdoor detection algorithm that can successfully detect and then remove the poison samples from a backdoor-injected ViT model with high success rates. The key intuition in our algorithm is that the patch processing strategies affect ViTs' predictions on the backdoor samples differently from the predictive function of the model on the clean data. Our defense algorithm exploits the frequency that ViTs change their predictions on the same sample under different trials of a patch processing strategy and use a threshold to assess if a sample is clean or poisoned. 
Our defense mechanism only requires access to a small set of $K$ clean samples (less than 1000 on the studied datasets), which can be easily obtained from the held-out validation dataset, for selecting the threshold. When no such clean samples are available, we show that the defenses are still very effective, which enables much wider applicability of the proposed method. The proposed defense consists of the following steps:

\newpage

\begin{itemize}
    \item \textbf{Step 1 (Offline):} For the small set of clean samples, randomly apply PatchDrop and PatchTranslate on each image for $T$ trials. For each sample $x$, we calculate $F_d(x) = \sum_{t=1}^{T} {\bf 1}\{F(x) \ne F(R_d^{(t)}(x))\}$ and $F_s(x) = \sum_{t=1}^{T} {\bf 1}\{F(x) \ne F(R_s^{(t)}(x))\}$, where $R_d^{(t)}$ and $R_s^{(t)}$ denotes the random application of PatchDrop and PatchShuffle, respectively, at trial $t$. Intuitively, $F_d(x)$ and $F_s(x)$ record the expected number of times the predicted labels on $x$ change to something else after the patch processing.
    \item \textbf{Step 2 (Offline):} Given the sample $\{F_d(x_i)\}_{i=1,..,K}$ or $\{F_s(x_i)\}_{i=1,..,K}$ created in Step 1, we set the threshold parameter $k_d$ and $k_s$ for PatchDrop and PatchShuffle, respectively, to the values at the $n^{th}$ percentiles to ensure a small false positive rate, as follows: 
    
    \begin{itemize}
        \item For PatchDrop, we typically select a large value (e.g., $90^{th}$ percentile). This is because ASRs significantly decrease under patch processing such as PatchDrop.
        \item For PatchShuffle, we typically select a small value (e.g., $10^{th}$ percentile). This is because ASRs do not drop under patch processing such as PatchShuffle while the clean-data accuracies are more affected. \vspace{-0.05in}
        
    \end{itemize}
    
    \item \textbf{Step 3 (During Inference):} For a sample in the inference stage, we randomly apply PatchDrop and PatchShuffle for $T$ trials and record the number of label changes, $F_d(x)$ and $F_s(x)$, respectively. If $F_d(x)$  is greater than the selected $k_d$ threshold for PatchDrop or  $F_s(x)$ is smaller than the selected $k_s$ threshold for PatchShuffle, we flag the sample as a backdoor sample. Otherwise, the $x$ is determined as a clean sample.

\end{itemize}


Note that, the proposed approach does not assume the knowledge of the type of backdoor attack, which ensures its practicality in various scenarios. Furthermore, when the model is benign, i.e., without the backdoor attack, because of the percentile selection rules, only a very small fraction of samples will be identified as false negatives. Formally, our defense approach follows a similar strategy as that of an anomaly detector. Thus, more sophisticated anomaly detection approaches can be used to improve the detection rate while keeping the false negative rate low; however, this is beyond the scope of this paper. The details of the detection algorithm are presented in Algorithm~\ref{code:detection}.\vspace{-0.04in}

    \begin{algorithm}[h]
    \caption{Patch Processing-based Backdoor Detection}
    \label{code:detection}
    \begin{algorithmic}[1]
    \Require Sample $x$, Threshold $k_d$ (PatchDrop),  Threshold $k_s$ (PatchShuffle)
    \Ensure Clean or Backdoor Decision
    \Function{F}{$x$}
        \State $t\leftarrow 0$, $F_d(x) \leftarrow 0$, $F_s(x) \leftarrow 0$, {Predict $\hat{y}=F(x)$}
        \Repeat 
            \State $t\leftarrow t+1$
            \State {$\hat{y}_{t} = F(R_t(x))$} 
            \State {$F_d(x) \leftarrow F_d(x)+1$ if $\hat{y}_{t} \ne \hat{y}$}
            \State {$\hat{y}_{t} = F(R_s(x))$} 
            \State {$F_s(x) \leftarrow F_s(x)+1$ if $\hat{y}_{t} \ne \hat{y}$}
        \Until $t = T$
        \State \Return {$F_d(x)$ and $F_s(x)$}
    \EndFunction
    \State {If $F_d(x) > k_d$ \\$\;\;\;\;$ or $F_s(x) < k_s$, $x$ is \textit{Backdoor}}
    \State  {Otherwise, $x$ is \textit{Clean}}
    \end{algorithmic}
    \end{algorithm}

\newpage

\subsection{Analysis of Patch Processing-based Defense}
We first provide a qualitative analysis of the proposed defense strategy for detecting both patch-based and blending-based backdoor samples from the backdoor-injected ViT and CNN models.

\subsubsection{Patch-based Attacks}

\begin{figure}[h]
\mbox{\hspace{-0.2in}
		\subfloat[CNN Models]{\includegraphics[width=6.9in]{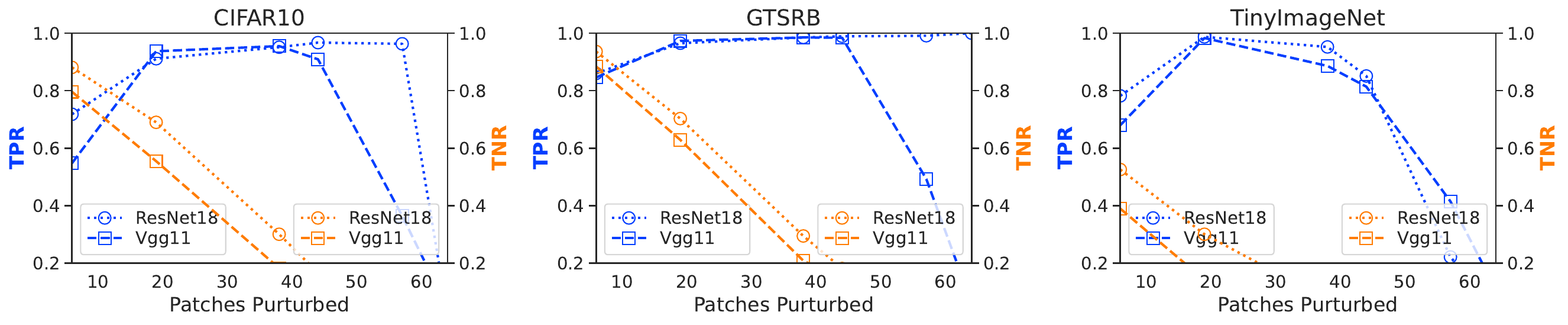}}
}

\mbox{\hspace{-0.2in}
		\subfloat[ViT Models]{\includegraphics[width=6.9in]{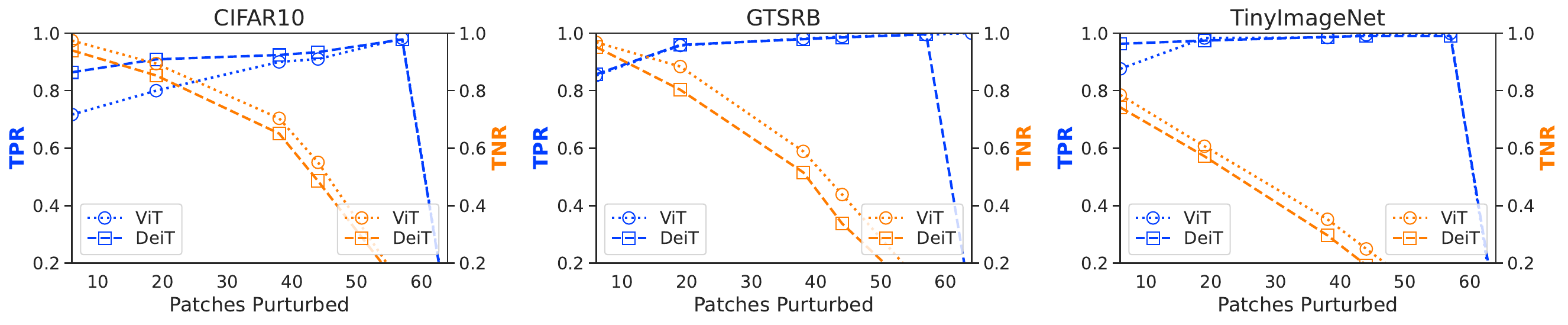}}
}

    \caption{TPR and TNR for different numbers of dropped patches (in a spatial grid of $8 \times 8$) for CNN models (ResNet18 and Vgg11) and ViTs (ViT and DeiT). 
    TPR: true positive rate, i.e., detection rate. TNR: true negative rate, i.e., clean-sample mis-detection rate.
    }
    \label{fig:result_drops_tpr}
\end{figure}

Figure~\ref{fig:result_drops_tpr} illustrates the true positive rate (TPR) and true negative rate (TNR) for the patch-based attack, BadNets, when varying the number of dropped patches $d$ in PatchDrop when the spatial grid is $8\times8$. Recall that the  TPR and TNR indicate the backdoor detection rate and the percentage of clean samples that are not falsely detected as backdoor samples, respectively. As we can observe, the defensive solution with PatchDrop works better for  ViT models than for CNN models such as ResNet18 and Vgg11. Furthermore, dropping 10\% of the patches can consistently achieve higher TPR and TNR across different datasets. The effectiveness of this defense on ViTs is because the backdoor performance is more sensitive to PatchDrop, as discussed in the previous section.

\newpage

\subsubsection{Blending-based Attacks}

\begin{figure}[h]
\mbox{\hspace{-0.3in}
		\subfloat[CIFAR10]{\includegraphics[width=0.35 \textwidth]{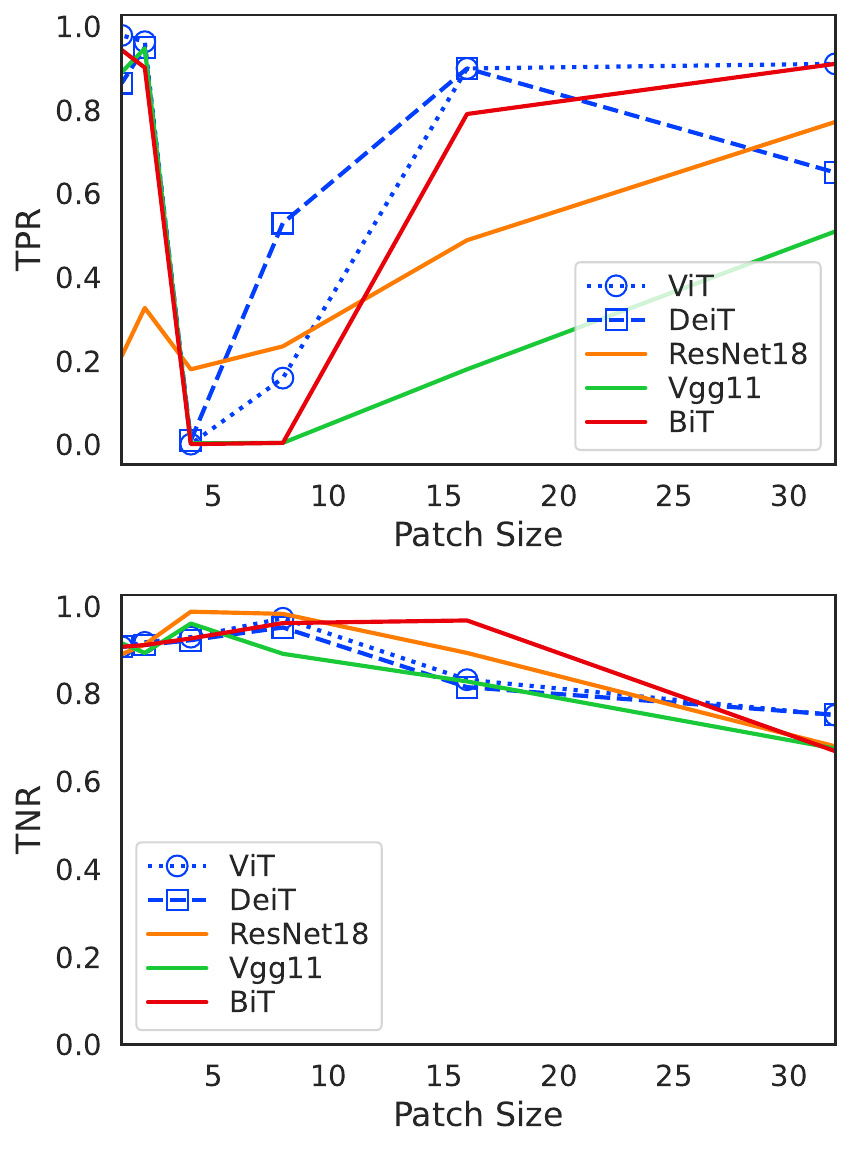}}
	\hfill
		\subfloat[GTSRB]{\includegraphics[width=0.35 \textwidth]{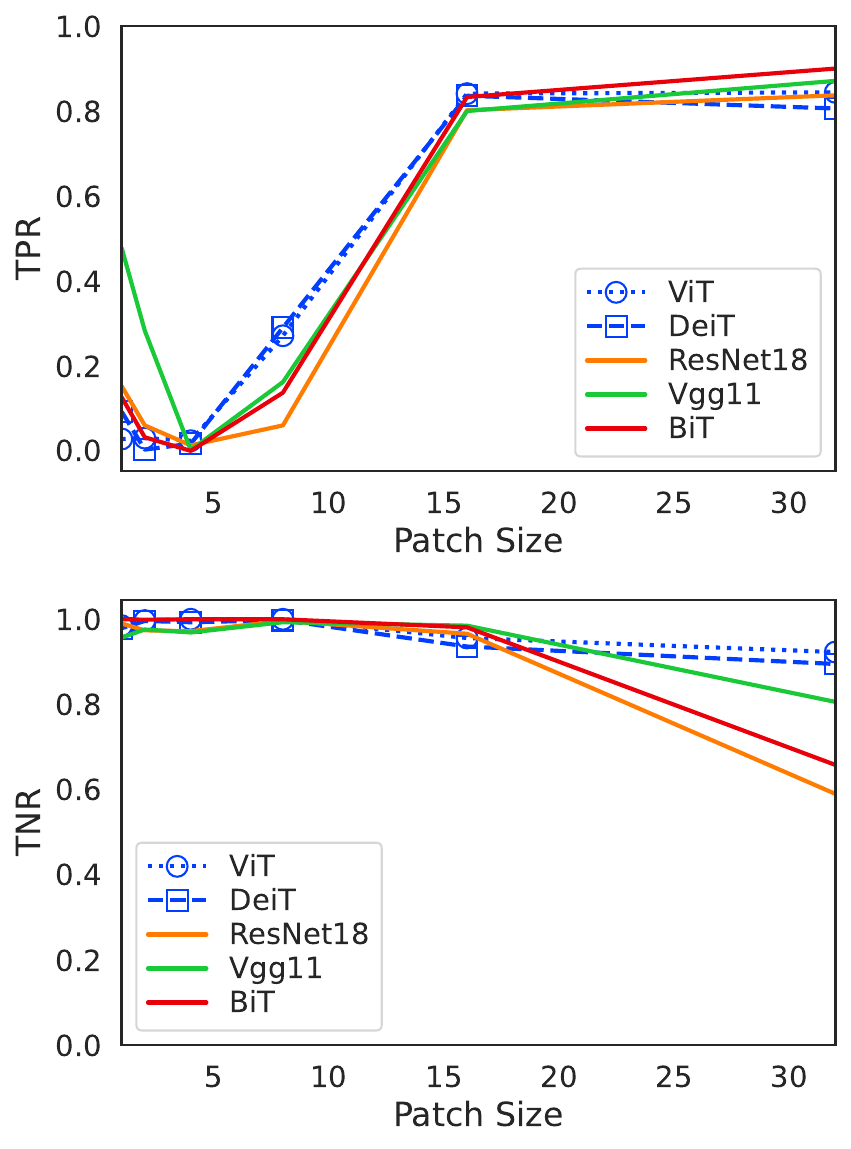}}
	\hfill
		\subfloat[TinyImageNet]{\includegraphics[width=0.35 \textwidth]{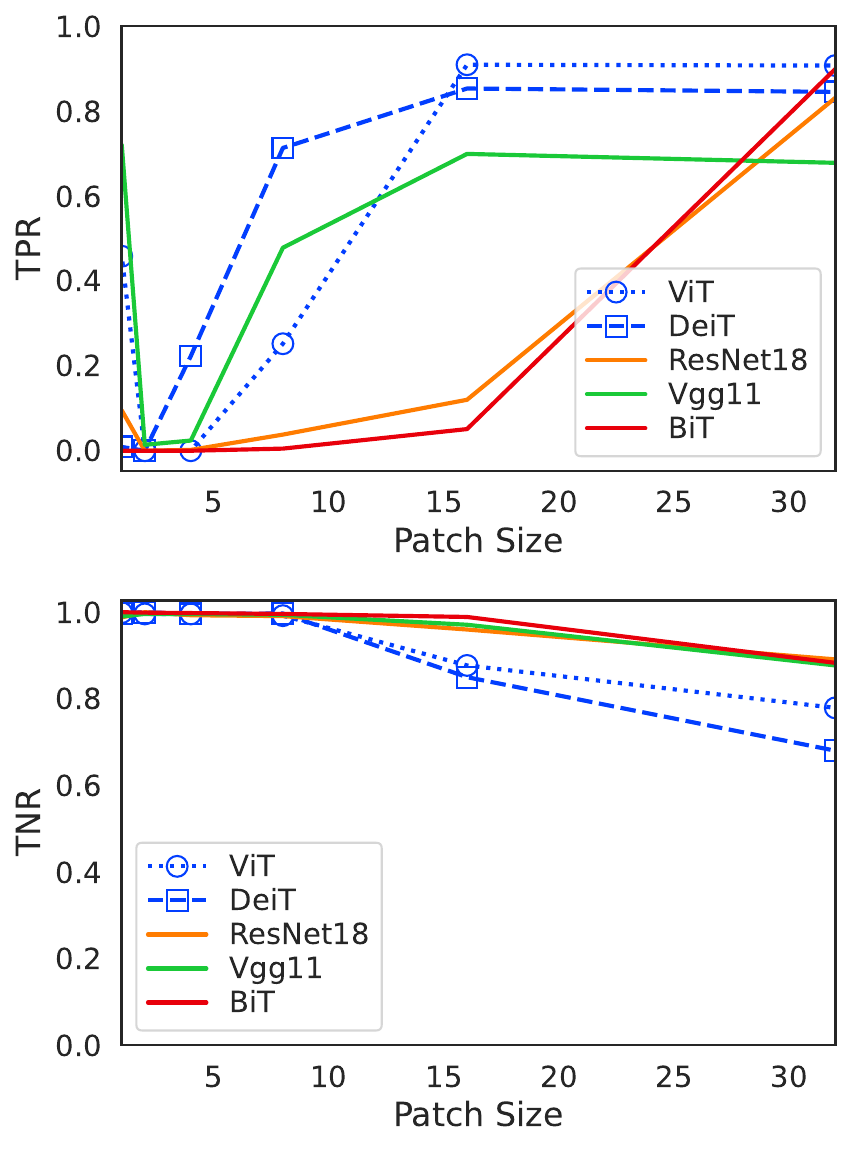}}
}		
    \caption{TPR and TNR for different sizes of processed patches for CNN models (ResNet18 and Vgg11) and ViTs (ViT and DeiT) under ReFool backdoor attack. }\vspace{0.2in}
    \label{fig:result_shuffle_tpr}
\end{figure}

Figure~\ref{fig:result_shuffle_tpr} illustrates the TPR and TNR when defending against ReFool with various sizes of the processed patches in PatchShuffle. As we can observe, PatchShuffle generally achieves higher TPRs in ViTs than in CNN models. More importantly, when the patch size is similar to that of the trained patch size in ViTs, defending against ViTs is consistently effective, with optimal TPRs and high TNRs. 

\section{Defense Experimental Results}

This section presents the empirical results in defending against backdoor attacks. In real-world settings, the defender does not know which attack is performed by the adversary. To this end, we consider two practical scenarios. 

In the first scenario, the backdoor is successfully injected into the trained model and the victim defends against backdoor attacks (i.e., alleviates its effectiveness) by filtering the backdoor samples during inference. In this experiment, TPR and TNR are reported, as they demonstrate how likely the defense method identifies the backdoor samples and how likely the clean samples are not falsely flagged as backdoor samples, respectively. We also assume that a small set of clean samples are available. The values at the $90^{th}$ and $10^{th}$ of the empirical distributions of $F_d(x)$ and $F_s(x)$, for all clean samples $x$, are selected as the thresholds $k_d$ and $k_s$ for PatchDrop and PatchShuffle, respectively.

\newpage

In the second scenario, we consider an extreme case where the defender is also the model trainer who receives a possibly poisoned training dataset. The defender aims to obtain the trained model that is free of the backdoor. Here, clean samples are not available, which makes the defending task very difficult. Defending using the proposed defensive solution consists of the following steps: 
\begin{enumerate}
    \item[(i)] train the model on the training dataset for some epochs, 
    \item[(ii)] use the possibly poisoned, trained model to detect the backdoor samples, and 
   \item[(iii)] remove the backdoor samples from the training dataset and re-train the model on the filtered dataset. 
 \end{enumerate}

We observe that only training the model for 50 epochs in step (i) is sufficient for the backdoor to be inserted into the model if the training dataset is poisoned and the clean-data accuracy reaches an acceptable performance compared to the optimal clean-data performance (less than a few percents difference, e.g., $>90\%$ in CIFAR10). Ideally, we train the models until they reach the optimal accuracies, but this can add significant computation to the training process while only adding a minor improvement in the defense. Therefore, we use 50 epochs on all experiments. In step (ii), we consider the threshold $k_d=0$ for PatchDrop, and $k_s=T$ for PatchShuffle.

\subsection{Defending against the Poisoned Model}

\begin{table}[ht!]
\newcommand{\change}[1]{\color{blue}#1\%}
\newcommand{\best}[1]{\textbf{\color{blue}#1\%}}
\newcommand{\second}[1]{\underline{\color{blue}#1\%}}
    \centering
    \caption{TPR and TNR of detecting backdoor samples in the poisoned models under BadNets Attack. Best TPRs are bolded. Best TNRs are underlined.}
    \begin{tabular}{l|c|c|c|c|c|c|c|c|c|c}
        \toprule
        Dataset & \multicolumn{2}{c|}{ViT} & \multicolumn{2}{c|}{DeiT} &
        \multicolumn{2}{c|}{Vgg11} & \multicolumn{2}{c|}{ResNet18}  & \multicolumn{2}{c}{BiT} \\ \cline{2-11}
        & TPR & TNR & TPR & TNR & TPR & TNR & TPR & TNR & TPR & TNR  \\ \hline
       CIFAR10 & 90.08 & \underline{99.48} & \textbf{91.88} & 96.18 & 88.12 & 62.88  & 88.80 & 89.33 &  90.00 & 87.72 \\ 
        GTSRB & \textbf{94.89} & \underline{98.78} & 93.62 & 97.66 & 20.15 & 80.70 & 93.80 & 89.99 &  93.89 & 92.91 \\ 
       TinyImageNet & 95.80 & \underline{64.75} & 95.80 & 64.73 & 81.30 & 20.51 & \textbf{99.00} & 56.48 &  98.60 & 42.64 \\

        \bottomrule
    \end{tabular}
    \label{tab:result_defense_experiments}
\end{table}

\begin{table}[ht!]
\newcommand{\change}[1]{\color{blue}#1\%}
\newcommand{\best}[1]{\textbf{\color{blue}#1\%}}
\newcommand{\second}[1]{\underline{\color{blue}#1\%}}
    \centering
    \caption{TPR and TNR of detecting backdoor samples in the poisoned models under ReFool Attack. Best TPRs are bolded. Best TNRs are underlined.}
    \begin{tabular}{l|c|c|c|c|c|c|c|c|c|c}
        \toprule
        Dataset & \multicolumn{2}{c|}{ViT} & \multicolumn{2}{c|}{DeiT} &
        \multicolumn{2}{c|}{Vgg11} & \multicolumn{2}{c|}{ResNet18}  & \multicolumn{2}{c}{BiT} \\ \cline{2-11}
        & TPR & TNR & TPR & TNR & TPR & TNR & TPR & TNR & TPR & TNR  \\ \hline
       CIFAR10 & \textbf{90.00} & 83.20 & 66.80 & 81.50 & 18.00 & 82.80 & 48.90 & \underline{89.30} & 79.10 & 66.90  \\ 
        GTSRB & \textbf{84.10} & 95.60 & 83.60 & 93.60 & 80.00 & 98.50 & 80.20 & \underline{96.60} & 83.30 & 92.10 \\ 
       TinyImageNet & \textbf{90.90} & 87.70 & 85.30 & 85.00 & 69.90 & 97.10 & 12.00 & 96.00 & 5.10 & \underline{98.90} \\
        \bottomrule
    \end{tabular}
    \label{tab:result_defense_experiments_refool}\vspace{0.3in}
\end{table}

Table~\ref{tab:result_defense_experiments} and Table~\ref{tab:result_defense_experiments_refool} present the defense results when the defender aims to detect whether a sample is a backdoor or clean sample during inference under the local patch-based attack, BadNets, and the global blending-based attack, ReFool, respectively.  As we can observe, the proposed defensive solution achieves comparable TPRs (i.e., successfully detects the backdoor samples) in both ViTs ($>$90\%) and CNN models ($>$88\%) across different datasets under BadNets attacks. However, the TNRs of ViTs, including ViT and DeiT, are significantly better than those of CNN models, including Vgg11, ResNet18, and BiT. Specifically, the proposed defense method only falsely detects clean samples as backdoor samples less than 3\% of the time in the trained ViTs, but more than 10\% of the time in the trained CNN models. Under ReFool attacks, the defensive solution achieves the best TPR for ViT, while its TNRs are also very high. While the TNRs of ResNet18 and BiT are higher than those of ViTs, their TPRs are significantly lower, especially in the larger-scale TinyImageNet dataset. Overall, we can conclude that the proposed defense method is consistently more effective in ViTs than in CNN models.

\subsection{Defending against the Poisoned Training Data}

\begin{table}[h!]
\newcommand{\change}[1]{\color{blue}#1\%}
\newcommand{\best}[1]{\textbf{\color{blue}#1\%}}
\newcommand{\second}[1]{\underline{\color{blue}#1\%}}
    \centering
    \caption{Clean-data accuracy and attack success rate after removing the backdoor samples and retraining the models. The second row (in \textcolor{blue}{blue}) in each dataset displays the relative change w.r.t the models trained without removing the backdoor samples. Best clean-data accuracy for each dataset is bolded.}
    \resizebox{\columnwidth}{!}{
    \begin{tabular}{l|c|c|c|c|c|c|c|c}
        \toprule
        Dataset & \multicolumn{2}{c|}{ViT} & \multicolumn{2}{c|}{DeiT} & \multicolumn{2}{c|}{ResNet18}  & \multicolumn{2}{c}{BiT} \\ \cline{2-9}
      & Clean & Attack & Clean & Attack & Clean & Attack & Clean & Attack  \\ \hline
      \multirow{2}{*}{CIFAR10} & \textbf{98.94} & 10.01 & 98.74 & 09.96 & 92.30 & 15.77 & 97.10 & 10.39 \\
      & \change{+0.01} & \change{--89.83} & \change{--0.08} & \change{--89.82}  & \change{--4.53} & \change{--88.99} & \change{--1.43} & \change{--89.30} \\ \cline{1-9}
       
       \multirow{2}{*}{GTSRB} & \textbf{98.38} & 0.48 & 97.98 & 0.47 & 96.58 & 0.46 & 96.89 & 0.48 \\
       &\change{--0.31} & \change{--99.51} & \change{--0.58} & \change{--99.51}  & \change{--2.31} & \change{--99.51} & \change{--1.85} & \change{--99.50} \\ \cline{1-9}
       
       \multirow{2}{*}{TinyImageNet} & 85.57 & 0.51 & \textbf{88.77} & 0.50  & 64.79 & 0.55 & 72.34 & 0.52 \\
       & \change{--1.03} & \change{--99.48} & \change{+1.15} & \change{--99.48} & \change{--5.65} & \change{--99.43} & \change{--10.68} & \change{--99.46} \\
        \bottomrule
    \end{tabular}
    }
    \label{tab:result_defense_retraining}\vspace{0.2in}
\end{table}

We present the clean-data accuracies and ASRs after re-training the models on the filtered data, as described in the second scenario, in Table~\ref{tab:result_defense_retraining}. The attack method is patch-based. We can observe that the proposed defense method successfully reduces the ASRs much closer to ASRs of random guesses in both ViTs and CNN models on all datasets. However, in ViTs, the clean-data accuracies are preserved, while in CNN models, the clean-data accuracies drop more than 4.5\% for ResNet18 and almost 1.5\% for BiT. The results for Vgg11 are worse than those of ResNet18 and BiT and are reported in supplement materials. As discussed in the previous experiment, a non-trivial number of clean samples can be falsely detected as backdoor samples in CNN models using the proposed patch-processing approach. Thus, while most backdoor samples are removed from the training datasets, the number of clean training samples is also reduced, which leads to the drop in clean-data performance in CNN models. We can also notice that by employing a large-scale pre-trained model (i.e., BiT), the drop in performance can be mitigated compared to smaller models, such as ResNet18 and Vgg11. Nevertheless, we can still observe that the proposed defense is more effective for ViTs than for CNN models. 

In conclusion, while ViT is vulnerable to patch-based backdoor attacks, our proposed simple yet effective patch processing based defensive solution can detect backdoor samples with a high detection rate while maintaining a low false negative rate. Because ViT is robust against patch processing on the clean data, processing the images with these strategies can be utilized to obtain useful yet tangible traces for effectively distinguishing the predictions between the clean and backdoor samples.

\section{Conclusion}

This paper studied several aspects of backdoor attacks against ViT. We first perform an empirical study on the vulnerability of ViT against both patch-based and blending-based backdoor attacks. Then, based upon our observation that ViT exhibits distinguishable performance between clean samples and backdoor samples against patch processing, we proposed a novel defensive solution to counter backdoor attacks on ViT, which is able to reduce the backdoor attack success rate significantly. Two patch processing methods are investigated. The effectiveness of the proposed techniques is comprehensively evaluated. To the best of our knowledge, this paper presented the first defensive strategy that utilizes a unique characteristic of ViT against backdoor attacks. 

\bibliographystyle{plainnat}
\bibliography{refs_scholar}
\end{document}